\documentclass[letterpaper, 10 pt, conference]{ieeeconf}
\usepackage{times}
\usepackage[pdftex]{graphicx}
\usepackage{}
\usepackage{amsmath,amssymb,amsopn,amstext,amsfonts}
\usepackage{pifont}
\usepackage{cancel}
\usepackage{cite}
\usepackage{pdfsync}
\usepackage{balance}
\usepackage{color}
\usepackage{mathtools}
\usepackage{algpseudocode}
\usepackage{textcomp}

\usepackage[ruled,vlined,linesnumbered]{algorithm2e}

\algnewcommand\algorithmicforeach{\textbf{for each}}
\algdef{S}[FOR]{ForEach}[1]{\algorithmicforeach\ #1\ \algorithmicdo}

\usepackage{bm}

\usepackage{diagbox}
\usepackage{pifont}
\usepackage{amsmath}
\usepackage{multirow}
\usepackage{url}
\usepackage{verbatim}
\usepackage{footmisc}
\usepackage[linkcolor=blue,citecolor=blue,urlcolor=blue,colorlinks=true]{hyperref}

\usepackage{color}
\usepackage{tabularx}
\usepackage{float}
\usepackage{threeparttable}
\usepackage{booktabs}
\usepackage{hyperref}
\graphicspath{{figures/}}
\DeclareGraphicsExtensions{.png,.jpg,.eps}
\IEEEoverridecommandlockouts
\newcommand{\MethodName}{Navi2Gaze\xspace}
\newcommand{\MyPara}[1]{\noindent\textbf{#1}}

\title{\MethodName: Leveraging Foundation Models for Navigation and Target Gazing}

\author{
	Jun Zhu$^{1,\ast}$, Zihao Du$^{1,\ast}$, Haotian Xu$^2$, Fengbo Lan$^1$, Zilong Zheng$^3$, Bo Ma$^4$,  \\ Shengjie Wang$^{5,\dagger}$, and Tao Zhang$^{1,\dagger}$,  \emph{Fellow, IEEE}
	\thanks{$\ast$
        denotes equal contribution.}
        \thanks{$\dagger$ denotes corresponding author: {\tt\small taozhang@tsinghua.edu.cn}, {\tt\small wangsj23@mail.tsinghua.edu.cn}.}
	\thanks{$^1$ Department of Automation, Tsinghua University.  $^2$Beijing Institute of Astronautical Systems Engineering. $^3$ Beijing Institute for General Artificial Intelligence. $^4$Chengdu Aircraft Design \& Research Institute. $^5$ Institute for Interdisciplinary Information Sciences (IIIS), Tsinghua University.}
}

\begin{document}
\maketitle
\newcommand\footnoteref[1]{\protected@xdef\@thefnmark{\ref{#1}}\@footnotemark}
\newlength{\bibitemsep}\setlength{\bibitemsep}{.0238\baselineskip}
\newlength{\bibparskip}\setlength{\bibparskip}{0pt}
\let\oldthebibliography\thebibliography
\renewcommand\thebibliography[1]{%
\oldthebibliography{#1}%
\setlength{\parskip}{\bibitemsep}%
\setlength{\itemsep}{\bibparskip}%
}
\vspace{-1.0cm}
\begin{abstract}
Task-aware navigation continues to be a challenging area of research, especially in scenarios involving open vocabulary. Previous studies primarily focus on finding suitable locations for task completion, often overlooking the importance of the robot's pose. However, the robot's orientation is crucial for successfully completing tasks because of how objects are arranged (e.g., to open a refrigerator door). Humans intuitively navigate to objects with the right orientation using semantics and common sense. For instance, when opening a refrigerator, we naturally stand in front of it rather than to the side. Recent advances suggest that Vision-Language Models (VLMs) can provide robots with similar common sense.
Therefore, we develop a VLM-driven method called Navigation-to-Gaze (\MethodName) for efficient navigation and object gazing based on task descriptions. This method uses the VLM to score and select the best pose from numerous candidates automatically. In evaluations on multiple photorealistic simulation benchmarks, \MethodName significantly outperforms existing approaches by precisely determining the optimal orientation relative to target objects, resulting in a 68.8\% reduction in Distance to Goal (DTG). Real-world video demonstrations can be found on the supplementary website\footnote{\href{https://sites.google.com/view/navi2gaze/}{https://sites.google.com/view/navi2gaze}}.
\end{abstract}

\section{Introduction}
The capability of zero-shot object navigation (ZSON) is deemed essential for the effective operation of home-assistance robots. Current research commonly views final locations close to targets as the measure of successful navigation. In practice, an appropriate orientation relative to the object, often overlooked by existing methods, is crucial for enhancing subsequent operations \cite{yenamandra2023homerobot, Saxena2009_object_orientation_from_images, he2023pick2place}. 
Consider an intuitive example: when a navigation system directs robots only to the refrigerator without addressing the need to open its door, the robots may end up positioned at an impractical side or back, thereby hindering the smooth execution of further actions. Similarly, for a task like sitting on a sofa, simply arriving nearby is inadequate; the robot must be properly aligned to face the sofa and positioned to sit down comfortably.
Although 6D object pose estimation can be obtained from recent work \cite{Sundermeyer_2018_ECCV, wada_2022_reorientbot, hai2023shape, wen2023bundlesdf, kokic2020learning}, the model still struggles to choose the appropriate orientation for specific tasks.

\begin{figure}[ht]
	\centering
	\includegraphics[width=0.5\textwidth]{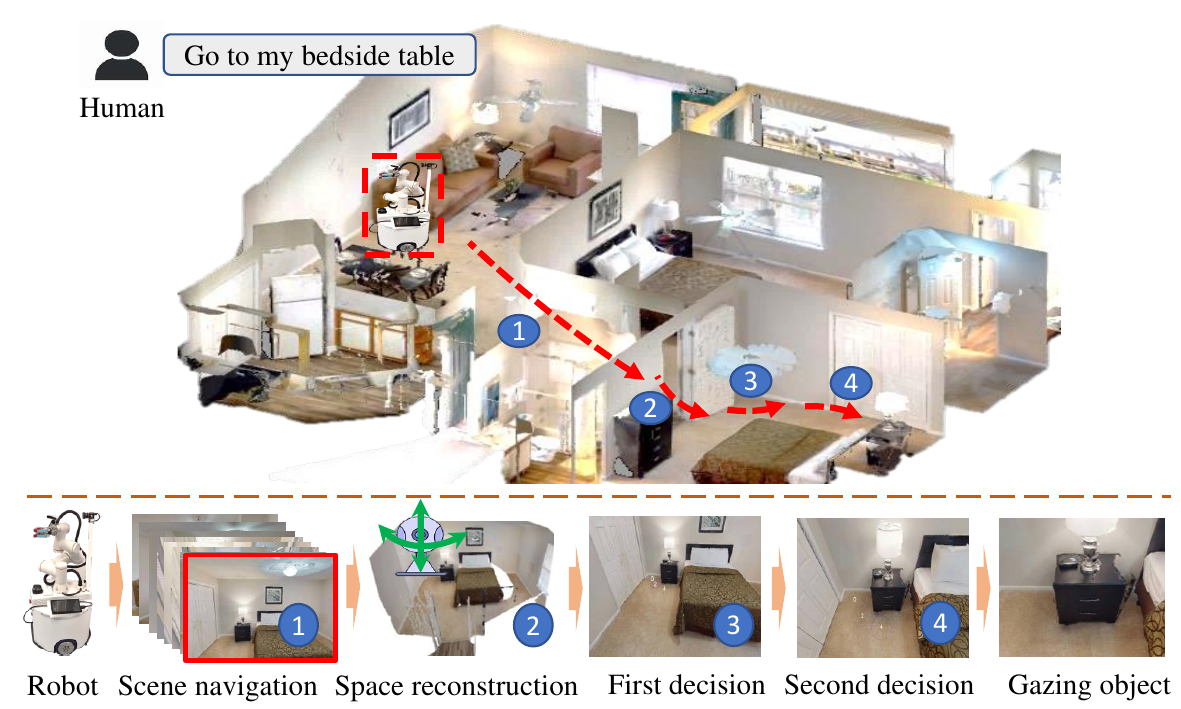}
	\vspace{-0.1in}
	\caption{\textbf{Workflow of \MethodName}: (1) Identification of target scene; (2) Reconstruction of task-aware space; (3) Generation of location candidates; (4) Sequential decisions for gazing objects.}
	\label{fig:001}
\end{figure}

Recently, Visual language models (VLMs) have emerged as a promising approach for understanding the world through images and text inputs, learning to represent these perceptions by projecting them into a linguistic space \cite{brown2020language,openai2024gpt4,chowdhery2022palm}. 
Due to the lack of spatial understanding, it is impractical for VLMs to directly generate navigation trajectories in text form.
Current state-of-the-art navigation algorithms\cite{majumdar2022zson, Huang2022VisualLM, cai2023bridging} predominantly utilize VLMs for common sense reasoning and image-level visual perception\cite{Li_2022_CVPR_GLIP, Radford_2021_CLIP}. These models process inputs ranging from original images to those augmented with colorful boxes or segmentation masks\cite{yang2023setofmark, Yang_2023_Fine_Grained_Visual_Prompting, yao2024cpt}. While these visual prompts enable VLMs to guide robots to the vicinity of the target object, they can not directly provide the precise orientation required for subsequent tasks, primarily due to the absence of 3D spatial projection information.
Thus, the question then arises: \textit{how can we use the common sense of VLMs to develop a task-aware navigation system that can navigate to the target object with the correct orientation?}

To address this challenge, we propose a VLM-driven Navigation-to-Gazing method, named \MethodName. Our method first provides potential candidates (target scenes, regions of the robot, and objects) to the VLM, then uses a coarse-to-fine mechanism to score candidates, and finally chooses the best pose for the robot. 
For instance, given the instruction ``open the refrigerator," VLMs can be prompted to:
1) Select the optimal scene where the refrigerator is visible,
2) Navigate to the region corresponding to this image, and
3) Identify the ideal position near the refrigerator that facilitates easy opening.
Navigation is executed by the underlying path planning algorithm, while VLMs function as the robot's decision-making brain. An illustrative diagram and a subset of tasks we considered are shown in Fig. \ref{fig:001}.
In summary, our contributions are threefold:
\begin{itemize}
	\item \textbf{Navigation with Gazing at the Object} - We propose a universal framework that navigates the robot toward the target object with the appropriate orientation, thus facilitating the success of further manipulation.
	\item \textbf{VLM-based Scoring Function} - To make full use of the common sense in VLMs, we design a VLM-based scoring function that automatically generates scores for the robot to choose the optimal pose, using visual observations with annotated candidates. We showcase its zero-shot generalization capabilities for open-set instructions involving open-set objects across a range of daily navigation tasks.
	\item \textbf{Navigation without a 3D Semantic Map} - Our method leverages a VLM-driven navigation policy to choose the best pose for the robot, given only a few visual observations and a simple 2D navigation map.
\end{itemize}

\section{Related Works}
\label{sec:relatedworks}
\subsection{Open-Vocabulary Object Navigation}
Most previous studies have employed contrastively trained visual language models (VLMs), such as GLIP\cite{Li_2022_CVPR_GLIP} and CLIP\cite{Radford_2021_CLIP}, to address the open-vocabulary object navigation challenge.
These models, trained on image-language associations, facilitate open-vocabulary image understanding and object detection.
For instance, ZSON \cite{majumdar2022zson} utilizes CLIP to map image-goals (e.g., a picture of the sink) and object-goals (e.g., "sink") into a semantic-goal embedding space. This approach facilitates training that enables agents to identify objects described in natural language.
An ensemble of ViLD\cite{gu2021open} and CLIP is utilized in NLMap\cite{Chen2022OpenvocabularyQS} to extract image embeddings that are queryable via text.
VLMaps\cite{Huang2022VisualLM} integrates pretrained visual-language features, extracted by LSeg\cite{li2022language}, with a 3D reconstruction of the physical world to generate a spatial map representation.
Recently, advancements in VLMs have demonstrated such robust vision-language understanding capabilities that they may potentially replace visual language models.
Furthermore, we have discovered that a few images can capture nearly all elements within a scene, allowing us to forego the creation of a semantic map for simplicity.

\subsection{Visual Prompting}

Despite impressive zero-shot transfer capabilities in image-level visual perception, recent studies\cite{yang2023setofmark, Yang_2023_Fine_Grained_Visual_Prompting, Shtedritski_2023_ICCV, subramanian2022reclip, yao2024cpt} indicate that both VLMs and large language models (LLMs) exhibit limited effectiveness in instance-level tasks requiring precise localization and recognition.
Therefore, researchers suggest that integrating visual prompts, such as colorful boxes, circles, precise semantic markers, and highlighted regions, can enhance the models' ability to recognize objects of interest. For example, SoM\cite{yang2023setofmark} employs SEEM\cite{NEURIPS2023_3ef61f7e} or SAM\cite{Kirillov_2023_ICCVSAM} to partition an image into regions with varying granularity, overlaying these regions with a variety of markers such as alphanumerics, masks, and boxes. Additionally, FGVP\cite{Yang_2023_Fine_Grained_Visual_Prompting} utilizes precise semantic masks of target instances derived via SAM. To generate fine-grained masks, the variants\cite{dai2023samaug, tang2023can}  of SAM, PerSAM\cite{zhang2023personalize}, and SAM-PT\cite{rajivc2023segment}, can also be  employed. Our approach employs visual prompts, similar to SoM, to select the optimal image in which the target object is visible. Besides, it draws colorful circles near the target on the image, enabling VLMs to identify the optimal position near the target object through a scoring mechanism.

\subsection{LLMs for Visual Navigation} 
The perceptual and decision-making capabilities of LLMs significantly enhance robotic navigation autonomy \cite{Driess2023PaLM_E, openai2024gpt4}.
For instance, L3MVN \cite{Yu_2023L3MVN} uses LLMs to identify relevant areas from a semantic map for better exploration. 
ESC \cite{Zhou2023ESC} leverages LLMs for commonsense reasoning about spatial relations between goal objects and common objects or rooms. 
LGX \cite{Dorbala2024Can_an_Embodied_Agent_Find} generates prompts from extracted object labels, guiding the navigation path. 
LFG \cite{shah2023lfg} uses positive and negative prompts for multiple samplings, providing scores as heuristics for the search algorithm. PixNav \cite{cai2023bridging} combines LLMs and VLMs to analyze panoramic images and determine room-level relationships, segment images, and guide robotic movements. 
StructNav \cite{chen2023train} uses language and scene-based priors to semantically score geometry-based boundaries, aiding in reasoning about promising search areas. 
LM-Nav \cite{shah2023lm} uses three pretrained models for planning: a large language model for landmark extraction, a vision-language model for localization, and a vision navigation model for execution.
We adopt the use of VLMs to improve object navigation efficiency. 
Unlike previous studies, we develop visual prompts that guide VLMs to accurately locate target objects and select the optimal orientation.

\section{Methodology}
\label{sec:methodology}

Leveraging the image recognition and natural language understanding capabilities of VLMs, a universal framework is introduced to guide robots towards the target object in the appropriate orientation. 
This framework effectively reduces the motion gap between navigation and manipulation, enhancing the efficiency and flexibility of robots operating in home environments. 

\begin{figure*}[ht]
	\centering
	\includegraphics[width=0.75\textwidth]{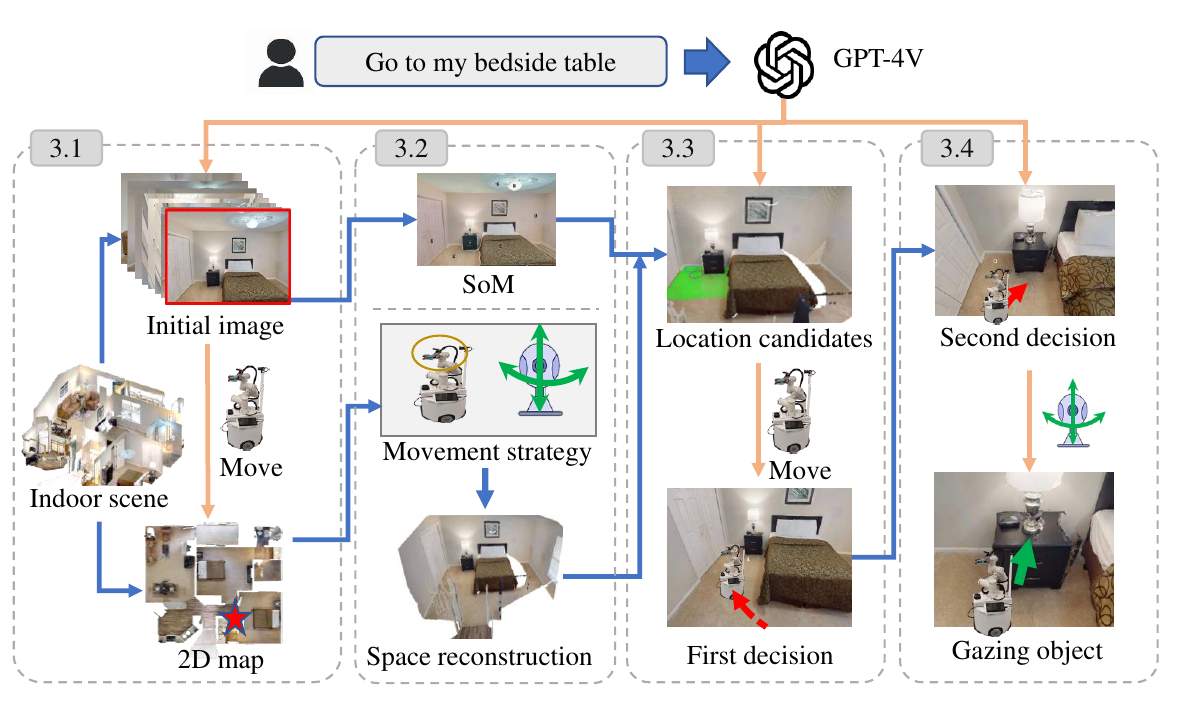}
	\vspace{-0.1in}
	\caption{\textbf{Framework of \MethodName}: 1. \textit{Identification of Target Scene}: The robot uses GPT-4V to find and navigate to the target object using image sequences. 2. \textit{Reconstruction of Task-aware Space}: The robot moves its camera to map the scene around the target object. 3. \textit{Generation of Location Candidates}: Self-organizing Maps (SoM) and GPT-4V identify the target and suggest possible regions for the robot to position itself. 4. \textit{Sequential Decisions for Gazing Objects}: GPT-4V scores these regions and directs the robot to the best position for observing the target.}
	\label{fig:framework}
	\vspace{-0.5em}
\end{figure*}

A graphical overview of the framework is shown in Figure \ref{fig:framework}. 
Our method processes a natural language query (e.g., ``open the refrigerator"), locates the queried object within a set $\mathcal{\boldsymbol{S}}=\{\boldsymbol{I}_i\}^N_1$ of $N$ images captured by the robot, navigates to the image location with the object, selects an optimal nearby position, and moves to this position.
We assume a 2D occupancy map $\mathcal{M}_o$ of the environment and an image set $\mathcal{\boldsymbol{S}}$ covering the scene. This is feasible in indoor settings where exploration can be done in one go.
We simplify the navigation task to a search problem: the robot proposes subgoals, evaluates them, and uses a search algorithm to plan the path. 
The core of \MethodName is scoring subgoal proposals from coarse to fine. 
First, it selects the image where the object is most visible, then it chooses the best nearby position to ensure optimal orientation.
\begin{figure}[htb]
	\centering
	\includegraphics[width=0.5\textwidth]{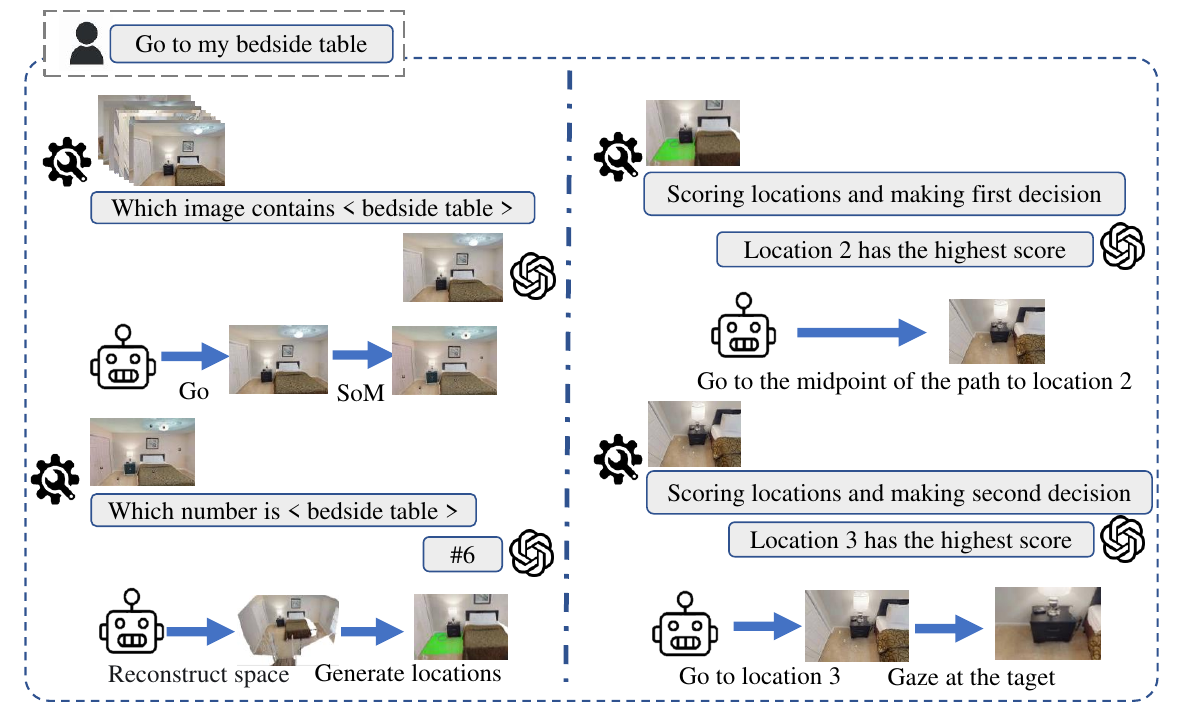}
	\vspace{-0.2in}
	\caption{Assistive process of GPT-4V in \MethodName. }
	\label{fig:004}
	\vspace{-0.5em} 
\end{figure}
The assistive process of GPT-4V in \MethodName is shown in Figure \ref{fig:004}.

\subsection{Identification of Target Scene}
\label{subsec:init_scoring}
Navigating to the photographic location of an image that includes the target object, using a set of scene images and a 2D occupancy map, constitutes the first critical step of our method. 
This strategy ensures that the target object appears within the robot's field of view. Specifically, our aim is to develop a scoring function utilizing a VLM that processes the image set $\mathcal{S}$ along with the goal query $q$ as inputs.
This function is designed to compute the task-relevant probability $p(\boldsymbol{I}_i, q, \mathcal{S})$ for each image $\boldsymbol{I}_i$ within $\mathcal{S}$.
To ensure precise indexing of VLMs, each image is marked with a unique red number in the upper left corner. 
We adopt a chain-of-thought prompting approach, which initially guides the VLM to identify images containing the queried object, then to discern the clearest image.
With the pose of the selected image determined, we project its 3D position onto a 2D occupancy map, thus establishing the 2D goal. 
This goal is then pursued using a 2D navigation system, effectively bridging the gap between image-based identification and physical navigation tasks.

\subsection{Reconstruction of Task-aware Space}
\label{sec:instantiation}
Obtaining extensive and up-to-date spatial information around the target object is the core prerequisite for guiding robotic navigation to the optimal orientation position. 
To this end, we divide the process into two components: target object recognition and extensive point cloud collection.

To accurately ascertain the location of the queried object, we employ the methodology used in SoM\cite{yang2023setofmark}. 
Initially, segmentation models such as SEEM\cite{NEURIPS2023_3ef61f7e} and SAM\cite{Kirillov_2023_ICCVSAM} are utilized to partition an image into regions based on semantic content. 
These regions are then distinctly colored and superimposed with various numeric markers.
Instead of relying solely on the marked image as in the SoM approach, we pair the original image with the marked, color-enhanced image before feeding them to the VLM. 
This dual-input strategy enhances the VLM’s ability to more effectively identify the queried object's numeric identifier in the image. 
Using this target number, we derive a sequence of precise coordinates for the target. 
These coordinates, in conjunction with the relevant depth data, are used to generate the 3D point cloud $\mathcal{P}_q$ of the target.

To optimally select a position that facilitates easy access to the target, we employ a systematic technique for collecting information about the area surrounding the queried object. 
Initially, the robot is positioned to directly face the target. 
It then rotates by an angle $\alpha_0= 30^{\circ}$ to both the left and right, during which it captures both image and depth data. 
The robot is subsequently tilted downward by a specified angle $\alpha_1=60^{\circ}$, and undergoes additional rotations of $\alpha_0$ to the left and right. 
This sequence enhances the collection of comprehensive image and depth data. Utilizing the accumulated images and depth data, the robot constructs a detailed point cloud $\mathcal{P}$ in the vicinity of the target.

\subsection{Generation of Location Candidates}
Numerous positions in proximity to the target object are accessible. 
Given the object's orientation attributes, such as the direction in which a refrigerator door opens or a drawer extends, only a subset of these positions are considered optimal for robotic manipulation. 
\begin{figure}[tb]
	\centering
	\includegraphics[width=0.5\textwidth]{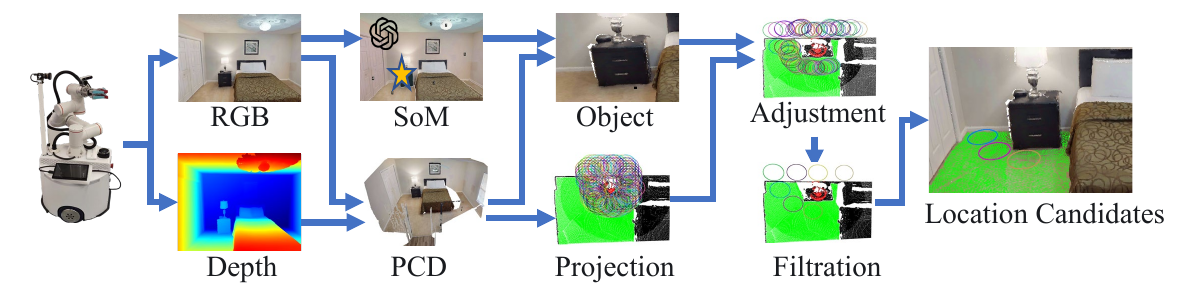}
	\vspace{-0.2in}
	\caption{Method for Generating Candidate Regions. }
	\label{fig:005}
	\vspace{-0.5em}        
\end{figure}
We propose a methodology for generating candidate regions, as shown in Figure \ref{fig:005}, thereby enabling more informed decision-making in subsequent phases.

Various point cloud data are projected onto the floor plane.
The ground's point cloud $\mathcal{P}_g$, can be extracted from $\mathcal{P}$ using a plane extraction algorithm RANSAC. 
Let $\boldsymbol{c}_q$ represent the center, and $r_q$ the radius, of the 2D projection of $\mathcal{P}_q$ on the ground.
On the ground, we construct a 2D square grid map $\mathcal{M}_s$ with an edge length of $r_q + 1$, a resolution of 1 cm, and centered at $\boldsymbol{c}_q$. 
We then project the non-ground point cloud, the ground point cloud $\mathcal{P}_g$, and the queried object point cloud $\mathcal{P}_q$ onto this grid map. 
Each grid can assume one of four possible values: $\emptyset$ signifies no projection point; $0$ indicates an obstacle point; $1$ represents an available ground point; $2$ denotes the queried object point (a special type of obstacle). 
The hierarchy of these values, from highest to lowest priority, is: $2$, $0$, $1$, $\emptyset$. 
Priority here implies replaceability—for instance, if a grid initially valued at $0$ receives a projection of $2$, it is updated to $2$; otherwise, it remains unchanged.

To ensure the robot is oriented correctly towards the target, we generate $K$ circular candidate regions $\{\mathcal{R}_i\}^K_{1}$, each with a radius $r_r$ (the robot's radius, which is 0.2 m in this case), around the 2D projection of the queried object on $\mathcal{M}_s$. We initiate by establishing a new grid on $\mathcal{M}_s$ at a resolution of $r_r/3$, utilizing the grid vertices as initial centers for the circles. Subsequently, circles are excluded if their minimum distance to the queried object points on $\mathcal{M}_s$ falls below $r_r/2$ or exceeds $3r_r/2$. We categorize different types of 2D points on $\mathcal{M}_s$ into distinct octrees\cite{zhu2023ioctree}, which facilitates the computation of distances. Following this, we pinpoint the nearest 2D projection point of the queried object to each of the remaining circles and ascertain the directional vector from the circle center to this nearest point. Each circle is then repositioned along this vector to ensure its separation from 2D obstacle points exceeds $r_r$ marginally. In the final phase, a circle is selected at random, and any overlapping circles are removed. This circle is retained as a candidate, and the process of selecting and retaining the closest non-overlapping circle continues until no circles are left. The circles that are ultimately retained fulfill the aforementioned criteria and are thus designated as the circular candidate regions.

\subsection{Sequential Decisions for Gazing objects}
Guiding the robot to the optimal region and orientation near the target is crucial for reducing the motion gap between robotic navigation and manipulation. 
To address this problem, we propose a two-step decision-making method designed to facilitate target gazing by robots. 
The candidate regions are projected onto the image plane with numeric markers, utilizing the cognitive capabilities of the VLM for analytical assessment. 
In the first step, we input the image containing candidate regions into the VLM for scoring to select the most promising circle. 
Subsequently, a path to the center of this circle is generated, and the robot is navigated to the midpoint of this path. 
In the second step, we re-project the candidate regions onto the current image, leveraging the VLM for a second analysis to select the final region. 
Upon arriving at the final region, the robot rotates to face the target object, thereby accomplishing target gazing.

\section{System Evaluation}
\label{sec:evaluation}
In this section, we conduct a comprehensive evaluation of our method through both simulation and real-world experiments, demonstrating its effectiveness in comparison to baseline methods.
Our experiments demonstrate that our method significantly outperforms existing LLM-based navigation algorithms, attributed to the integration of high-quality scores used as navigation heuristics.

\subsection{Simulation Setup}
\MyPara{Experimental setup:}
We conduct our simulation experiments using the Habitat simulator \cite{Savva_2019_ICCV_Habitat} and the large-scale indoor dataset HM3D v0.2 \cite{ramakrishnan2021hm3d}.
We collected RGB-D frames from nine different scenes, using a camera height of 1.5 meters and image resolution of $1080\times 720$. 
Additionally, we documented the camera pose for each frame. 
We evaluate the performance of \MethodName on the task of object-goal navigation, leveraging insights from the Habitat ObjectNav Challenge~\cite{habitatchallenge2022}. In this challenge, the agent operates within a simulated environment featuring photorealistic graphics and is tasked with locating a queried object from one of ten categories, such as ``toilet," ``bed," and ``couch."
In these environments, the robot is required to navigate in a continuous environment with actions: $\{stop, move\_forward, turn\_left, turn\_right, look\_up, \\look\_down\}$.
The forward distance is set to 0.1m, and the rotation angle is set to 1 degrees.
We evaluate our method and the baselines on a set of 68 tasks across these scenes.

\begin{figure}[htb]
	\centering
	\setlength{\abovecaptionskip}{0.cm}
	\includegraphics[width=0.4\textwidth]{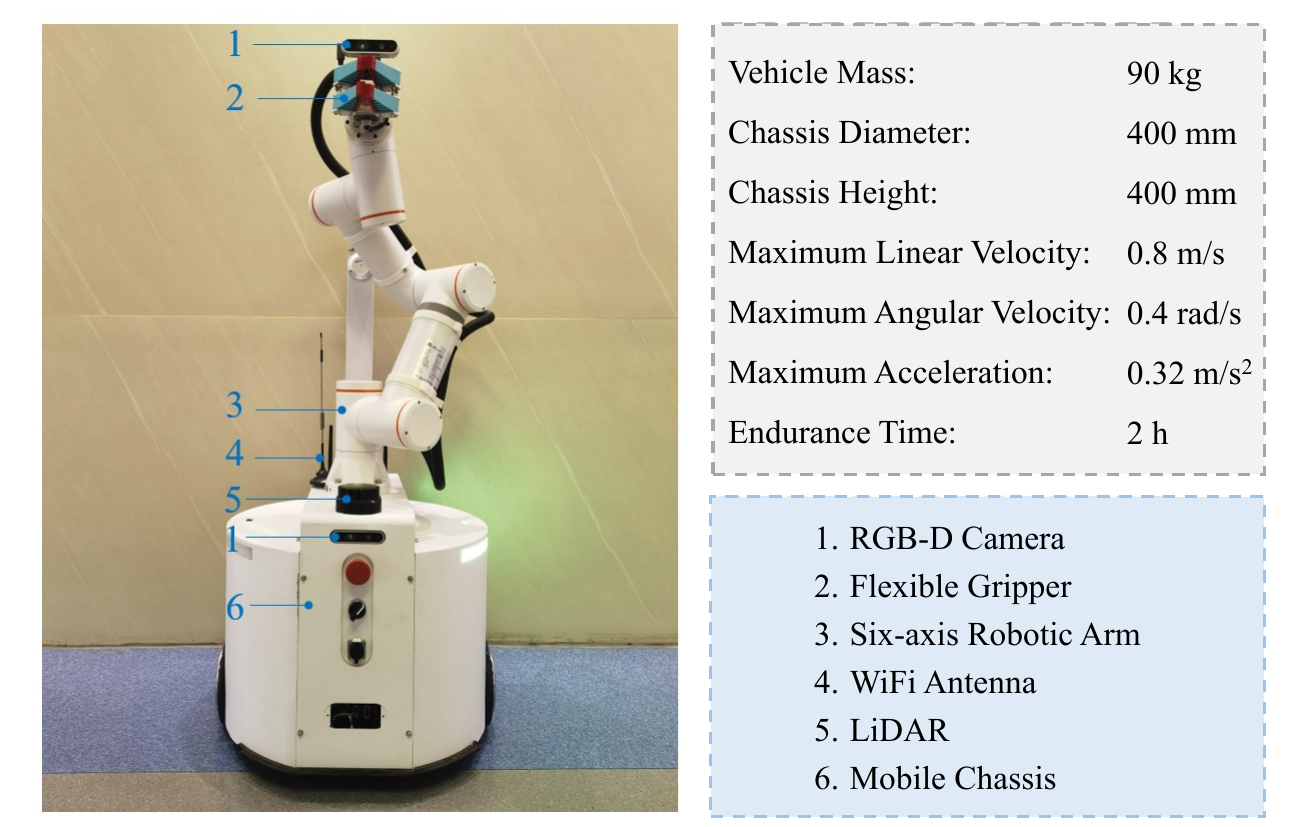}
	\vspace{-0.1in}
	\caption{Mobile manipulation robot. }
	\label{fig:robot}
	\vspace{-1.5em}
\end{figure}

We conducted real-world experiments using a custom-designed and manufactured robot, with the various components illustrated in Figure \ref{fig:robot}. 
The RGB-D camera used is an Intel RealSense Depth Camera D435i, with a resolution of 480x640. The experiments were carried out in our laboratory, using common items as target objects, such as a drawer, water dispenser, chair, fridge, and vending machine. 

\MyPara{Evaluation Metrics:}
We follow \cite{anderson2018evaluation} to evaluate our method using SR, Success weighted by Path Length (SPL), and DTG. 
\begin{figure}[htb]
	\centering
	\setlength{\abovecaptionskip}{0.cm}
	\includegraphics[width=0.4\textwidth]{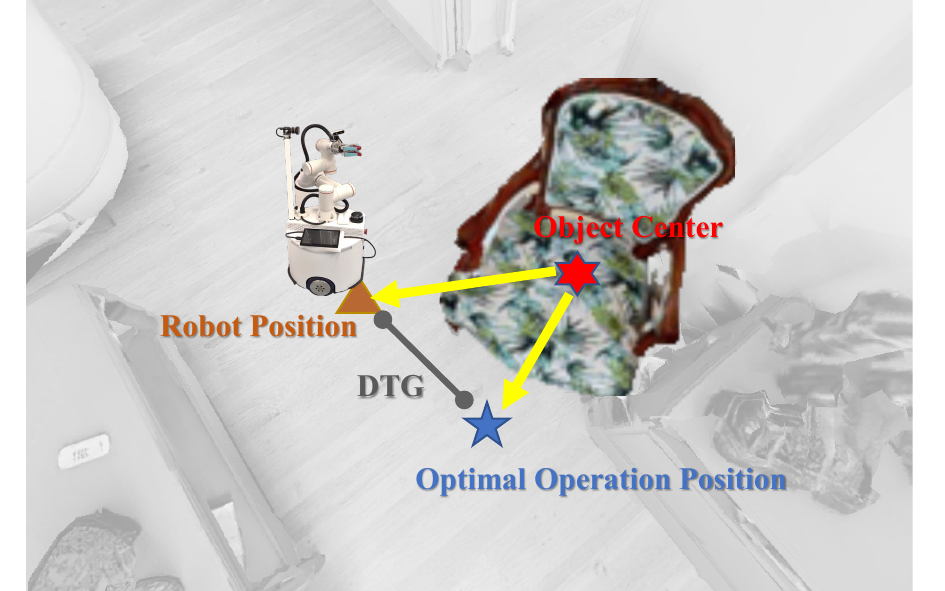}
	\vspace{0.0in}
	\caption{Distance to Goal (DTG): the distance between the Robot Position and the
Optimal Operation Position at the conclusion of the episode. }
	\label{fig:otg_dtg}
	\vspace{-0.5em}
\end{figure}
SR is defined as $\frac{1}{N}\sum_{i=1}^{N}S_i$ and SPL is defined as $\frac{1}{N}\sum_{i=1}^{N}S_i\frac{l_i}{max(l_i,p_i)}$ , where $N$ is the number of episodes, $S_i$ is binary indicator of success in episode $i$, $l_i$ is the shortest path length between the start position and the goal, $p_i$ is the path length of the current episode $i$. 
Specifically, we consider a case to be successful when the distance DTG is less than 0.5 m.
DTG represents the separation between the agent and the goal at the conclusion of the episode. 
Given our focus on ensuring that the robot not only navigates to the proximity of the object but also aligns itself with the correct orientation towards the object, we have redefined the SR and DTG in Figure \ref{fig:otg_dtg}. 

\MyPara{Baselines:}
We evaluate \MethodName against three baseline methods: VLMaps\cite{Huang2022VisualLM}, L3MVN~\cite{Yu_2023L3MVN} , and SemExp\cite{ChaplotNEURIPS2020SemExp}.
By combining GPT-3 and CLIP, VLMaps translates natural language instructions into a sequence of open-vocabulary goals, directly localized in the map. 
L3MVN builds the environment map and selects the long-term goal based on the frontiers with the inference of VLMs to achieve	efficient exploration and searching. 
SemExp builds an episodic semantic map and uses it to explore the environment efficiently based on the goal object category.
\subsection{Comparison Experiment}
\begin{figure}[tb]
	\centering
	\setlength{\abovecaptionskip}{0.cm}
	\includegraphics[width=0.5\textwidth]{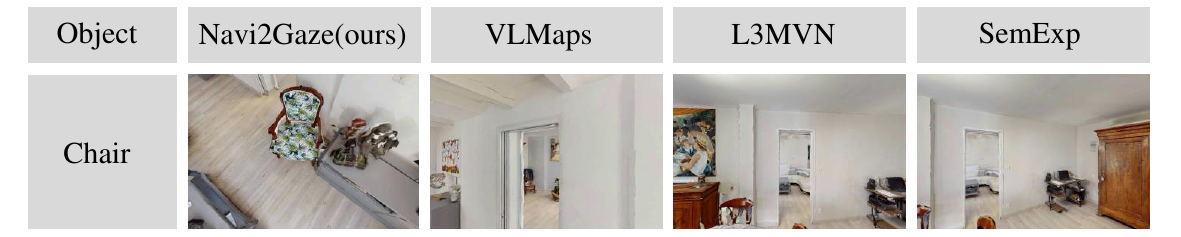}
	\vspace{-0.1in}
	\caption{The visualization of final observations from four navigation methods targeting a chair reveals that \MethodName effectively directs the robot to the front of the chair. Conversely, other methods frequently result in the robot approaching the chair from the back or side. }
	\label{fig:final_state}
	\vspace{-4.0em}
\end{figure}

We have designed two types of in-room experiments to evaluate the performance of object navigation. 
\begin{table*}[h]
	\centering
	\caption{Performance Comparison with all baseline}
	\resizebox{0.85\textwidth}{!}{%
		\begin{tabular}{@{}lllllllllllll@{}}
			\toprule
			\multicolumn{1}{c}{\multirow{2}{*}{Method}} & \multicolumn{2}{c}{Scene A} & \multicolumn{2}{c}{Scene B} & \multicolumn{2}{c}{Scene C} & \multicolumn{2}{c}{Scene D} & \multicolumn{2}{c}{Scene E} & \multicolumn{2}{c}{Scene F} \\
			\cmidrule(l){2-3} \cmidrule(l){4-5} \cmidrule(l){6-7}  \cmidrule(l){8-9} \cmidrule(l){10-11} \cmidrule(l){12-13}
			\multicolumn{1}{c}{} & SR & DTG  & SR & DTG  & SR & DTG  & SR & DTG  & SR & DTG  & SR & DTG  \\ 
			\midrule
			SemExp & 0.06 & 0.62 & 0.04 & 0.53  & 0.06 & 0.81  & 0.10 & 0.67  & 0.00 & 1.00  & 0.02 & 1.14  \\
			L3MVN & 0.14 & 0.87  & 0.20 & 0.39  & 0.19 & 0.61  & 0.19 & 0.66  & 0.00 & 0.87  & 0.09 & 2.27  \\
			VLMaps & 0.25 & 1.77  & \textbf{0.57} & 0.57  & 0.36 & 1.81  & 0.42 & 2.08  & 0.00 & 1.05  & \textbf{0.63} & 0.78  \\
			Navi2Gaze & \textbf{0.56} & \textbf{0.26}  & 0.55 & \textbf{0.26}   & \textbf{0.95} & \textbf{0.19}  & \textbf{0.71} & \textbf{0.34}  & \textbf{0.60} & \textbf{0.34}  & 0.60 & \textbf{0.66}  \\
			\bottomrule
		\end{tabular}%
	}
	
	\label{tab:all_baselines}
\end{table*}
In the first experiment, the initial positions of the robot are randomly selected, and all objects available in the scenes are used for the baselines. 
\begin{table}[ht]
	\centering
	\caption{Comparing Performance of VLMaps with Initial Poses on Object Sides}
	\resizebox{0.5\textwidth}{!}{%
		\begin{tabular}{@{}llllllllll@{}}
			\toprule
			\multicolumn{1}{c}{\multirow{2}{*}{Method}} & \multicolumn{3}{c}{Scene G} & \multicolumn{3}{c}{Scene H} & \multicolumn{3}{c}{Scene I} \\
			\cmidrule(l){2-4} \cmidrule(l){5-7} \cmidrule(l){8-10}
			\multicolumn{1}{c}{} & SR & SPL & DTG  & SR & SPL & DTG  & SR & SPL & DTG  \\ \midrule
			VLMaps & 0.65 & \textbf{0.65} & 0.54  & 0.25 & 0.25 & 0.80  & 0.23 & 0.23 & 1.79  \\
			Navi2Gaze & \textbf{0.68} & 0.46 & \textbf{0.37}  & \textbf{0.36} & \textbf{0.30} & \textbf{0.51}  & \textbf{0.43} & \textbf{0.26} & \textbf{0.56}  \\
			\bottomrule
		\end{tabular}%
	}
	\label{tab:all_baselines2}
 \vspace{-1.3em}
\end{table}
The outcomes of these experiments are presented in Table~\ref{tab:all_baselines}. 
Our observations indicate that \MethodName consistently outperforms all baselines. 
Both SemExp and L3MVN frequently get stuck due to their improperly selected short-term goals, particularly when the subsequent actions prescribed by these goals are unachievable. Although VLMap relies on an occupancy map, the RGB-D frames employed to construct this map may not capture all obstacles, leaving some areas unseen. Consequently, the path planning algorithm of VLMap treats these unseen areas as passable, often resulting in erroneous paths toward the target objects. Additionally, the robot frequently reaches the back or side of objects, positions that are typically suboptimal. An example of the final observations is illustrated in Figure \ref{fig:final_state}.

In the second experiment, we select the initial poses of the robots at the side or back of the objects. 
This configuration demonstrates that our method effectively guides the robots to the correct positions with the appropriate orientation, thereby validating the efficacy of our visual prompting and scoring mechanisms.
We mainly compare \MethodName with VLMaps in these experiments and the results are shown in Table~\ref{tab:all_baselines2}.
Our findings indicate that with appropriate visual prompts, VLMs can effectively determine the optimal positions for the robots, resulting in our method significantly outperforming VLMaps across all metrics.

By combining the results from the first and second experiments, our method demonstrated an average 61.9\% increase in SR compared to VLMaps, along with an average 68.8\% reduction in OTG. These experimental results validate the effectiveness of our approach.
\subsection{Ablation Studies}
To assess the relative importance of the various modules within our framework, we have performed the ablations using the HM3D dataset: Directly Navigating to the Target (\MethodName-DNT), without Reconstruction of Task-aware Space (\MethodName-w/o RTS), One-step Goal Decision (\MethodName-OGD). 
\begin{table}[htb]
	\centering
	\caption{ablation studies}
	\resizebox{0.4\textwidth}{!}{%
		\begin{tabular}{lccccccccccc}
			\hline
			\multicolumn{1}{c}{} & SR & SPL & DTG  \\ \hline
			\MethodName & \textbf{0.43} & \textbf{0.26} & \textbf{0.56}  \\
			\MethodName-DNT &  0.02 & 0.01 & 1.24  \\
			\MethodName-OGD & 0.32 & 0.24 & 0.85  \\
			\MethodName-w/o RTS & 0.15 & 0.11 & 1.22 \\
			\hline
		\end{tabular}%
	}
	\label{tab:Ablation}
 \vspace{-1.5em}
\end{table}
The results in TABLE \ref{tab:Ablation} show that our complete framework achieves the optimal performance, and the scoring mechanism and sequential decisions are crucial for achieving good performance. 
Our findings indicate that direct navigation to the target often fails, resulting in a significantly larger DTG. 
Without a reconstruction of task-aware space, the robot struggles to accurately recognize the ground due to insufficient information. Furthermore, replacing sequential decisions with a single-step goal decision leads to a decrease in both SR and SPL.

\subsection{Real-world Experiments}
\begin{figure*}[htb]
	\centering
	\setlength{\abovecaptionskip}{0.cm}
	\includegraphics[width=0.9\textwidth]{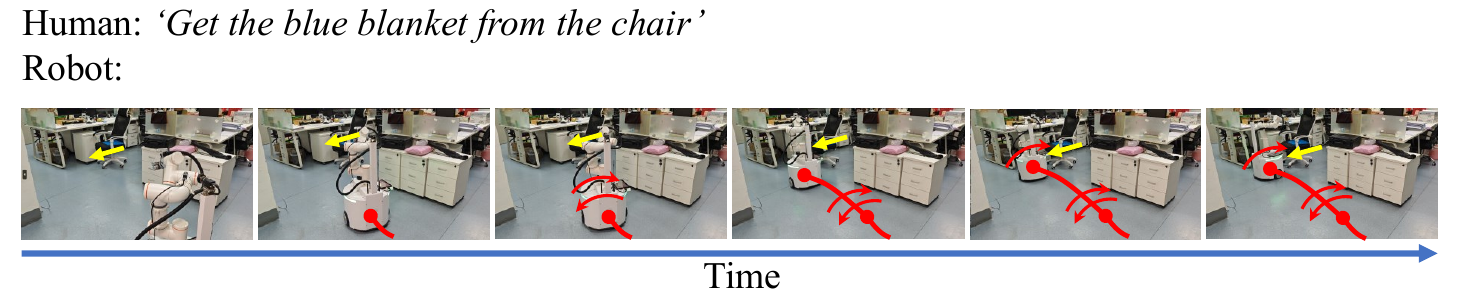}
	\caption{Movement process of mobile manipulation robot. The yellow arrow indicates the operational direction of the object. The red curves represent the planned path the robot follows while navigating through the space, and the red arrows signify left and right rotations.}
	\label{fig:demo}
	\vspace{-0.5em}
\end{figure*}
For the real-world experiments, the robot's movement process is illustrated in Figure \ref{fig:demo}. 
The yellow arrow indicates the operational direction of the object. 
The red curves represent the planned path the robot follows while navigating through the space, and the red arrows signify left and right rotations.
\begin{figure}[htb]
	\centering
	\setlength{\abovecaptionskip}{0.cm}
	\includegraphics[width=0.45\textwidth]{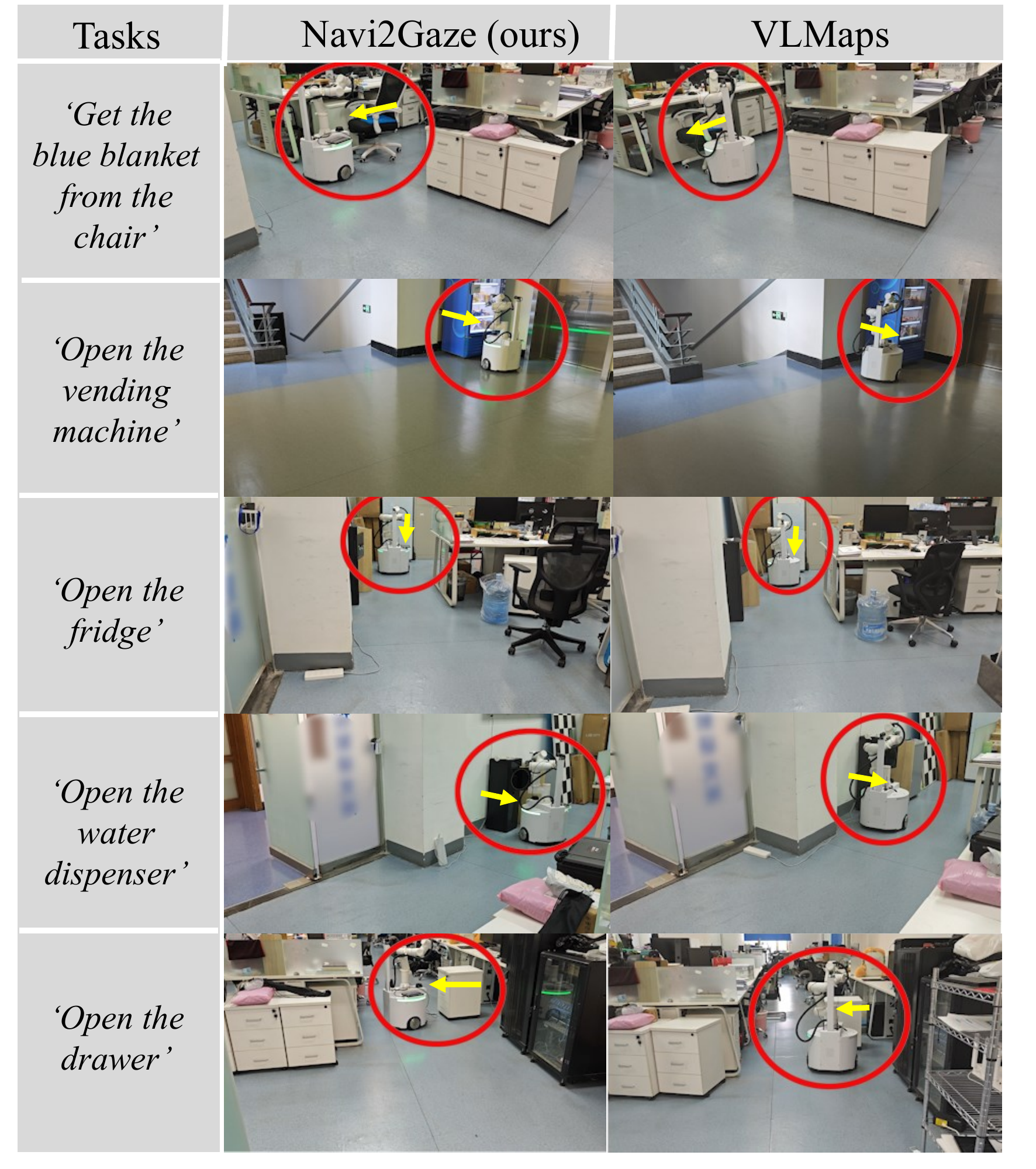}
	\vspace{-0.1in}
	\caption{Comparison with baseline. }
	\label{fig:tasks}
	\vspace{-0.5em}
\end{figure}
The results demonstrate the feasibility and effectiveness of our method, as shown in Figure \ref{fig:tasks}.
The comparison across tasks suggests that \MethodName performs better in terms of precise navigation and positioning, especially when approaching specific objects like the fridge, water dispenser, and drawer. 
Due to its inability to infer the optimal position for a given task, VLMaps can only approximate the proximity of objects, resulting in minor positioning inaccuracies that may impact task completion.
Overall, \MethodName demonstrate more consistent task execution and alignment with objects, indicating its potential effectiveness in real-world experiments.

\section{Conclusion \& Limitations}
\label{sec:conclusion}
We present \MethodName, a method using VLMs to help robots achieve precise goals. Our key finding is that VLMs provide valuable semantic insights but should guide navigation as heuristics, not fixed plans. We've created a way to derive heuristic scores from VLMs using visual prompts, and we incorporate these into our navigation algorithm. This approach improves navigation, helping robots accurately reach their desired positions and orientations.

Despite its potential, several limitations remain. 
First, the effectiveness of visual prompting hinges on precise image partitioning; however, the parameters of segmentation models can fluctuate under varying lighting conditions. Second, \MethodName depends on an accurate point cloud to correctly segment targets, ground, and obstacles—a task complicated by the often imprecise depths provided by RGB-D cameras.


\bibliography{navi2gaze}

\end{document}